\documentclass[sigconf]{aamas}  

\usepackage{booktabs}
\newtheorem{observation}{Observation}
\usepackage{comment}
\usepackage{color, colortbl}
\usepackage{booktabs}
\definecolor{Green}{RGB}{136,255,148}
\usepackage{flushend}
\setcopyright{ifaamas}  
\acmDOI{doi}  
\acmISBN{}  
\acmConference[AAMAS'20]{Proc.\@ of the 19th International Conference on Autonomous Agents and Multiagent Systems (AAMAS 2020), B.~An, N.~Yorke-Smith, A.~El~Fallah~Seghrouchni, G.~Sukthankar (eds.)}{May 2020}{Auckland, New Zealand}  
\acmYear{2020}  
\copyrightyear{2020}  
\acmPrice{}  
\renewcommand\footnotetextcopyrightpermission[1]{} 
\begin{document}

\title{Agent-Based User-Adaptive Filtering for Categorized Harassing Communication}  




%

%
\author{Zenefa Rahaman}
\affiliation{
\institution{The University of Tulsa}
\city{Tulsa, Oklahoma} }
\email{zenefa.rahaman@gmail.com}

\author{Sandip Sen}
\affiliation{
\institution{The University of Tulsa}
\city{Tulsa, Oklahoma} }
\email{sandip@utulsa.edu}
%
%
%
%

\begin{abstract}
Online communication platforms face persistent challenges in detecting and mitigating harassing communication while preserving contextual nuance and individual differences in tolerance. Traditional moderation systems typically rely on globally trained classifiers, implicitly assuming uniform user perception of harassment. However, psychological and behavioral research suggests substantial variability in how individuals perceive and respond to different categories and intensities of aggressive communication.

In this work, we propose an agent-based, user-adaptive filtering architecture for categorized harassing communication. We introduce a taxonomy of harassment types grounded in interdisciplinary literature and construct a labeled Twitter dataset augmented with large-scale crowdsourced user evaluations (over 26,000 responses). Statistical analyses, including ANOVA, Tukey HSD, and Wilcoxon tests, demonstrate significant variability in filtering preferences across users, harassment categories, and perceived intensity levels.

Motivated by these findings, we develop personalized, agent-based filtering mechanisms trained on individual user feedback and compare their performance to a general population-level classifier and a user-specific majority baseline. Experimental results show that user-adaptive filters consistently outperform general models in aligning with individual filtering preferences.

This work highlights the importance of personalization-aware intervention mechanisms in online safety systems and provides an early architectural framework for adaptive moderation in social media environments.

\textit{
 This paper is a revised and archived version of work originally presented at the Adaptive and Learning Agents Workshop (ALA 2019), co-located with AAMAS 2019. Minor revisions have been made for clarity and archival purposes.}
\end{abstract}

%


\maketitle


\section{Introduction}
In the current culture of widespread social media usage and high participation in online communication, the number of users and the time spent online per user is on the upswing. According to a report by the Pew Research Center, nearly 76\% of American adults use social networking sites in 2017, which was only 7\% in 2005~\cite{pew2017}. Online communication and information-sharing platforms, which compose the social media space, have empowered people to express their views and share their opinions. Social media users write about their lives, share opinions on a variety of topics, and discuss current issues. These sites have become the representative collections of people's opinions and sentiments as more and more users post about services they use, or express their political and religious views. 
Curated repositories of social media data serve as valuable data sources for businesses, researchers, and policymakers. Whereas these new communication channels, such as online social networks~\cite{kaplan2010users} and news sharing sites~\cite{lee2012news}, offer myriad opportunities for knowledge sharing and opinion mobilization~\cite{boyd2010tweet}, they also reveal an abundance of unfortunate intimidatory and hateful aggression~\cite{o2011impact} towards individuals targeted~\cite{willard2006cyberbullying} because of their expressed opinions or identities. 

Cyber harassment is an aggressive and unwanted online behavior that involves intimidation and victimization of targeted users with aggressive, derogatory, intimidatory, or demeaning language and/or visual communication.
Such behavior is typically frequent and repeated over time, but can also occur as an isolated incident.
A recent study~\cite{duggan2014online} found that 40\% of adult Internet users have experienced online harassment, with young women enduring particularly severe forms of such unwanted and undesirable communication. 38\% of women who had been harassed online reported the experience could be described as extremely upsetting. These nasty, and at times coordinated, victimization of individuals~\cite{campbell2005cyber} have significant social costs varying from social ostracism to opinion marginalization and suppression, and can cause severe health detriments ranging from anxiety~\cite{shaw2002defense} to depression~\cite{ybarra2004linkages} to suicide ideation~\cite{hinduja2010bullying,paul2002suicide}. 

Victims of such online attacks are often minority groups or individuals voicing dissent, professionals covering controversial topics essential for informing a democratic society, and groups raising awareness about important issues~\cite{bernstein2014abuse}.
A study of offline and online harassment of female journalists found that two-thirds of the respondents reported acts of intimidation, threats, and abuse related to their work~\cite{barton2014violence}.

While several computational studies have developed automated mechanisms for the detection of unwarranted victimization and harassing attacks on social network and micro-blogging platforms~\cite{nobata2016abusive,yin2009detection,sood2012using},
more comprehensive detection and intervention mechanisms that are grounded in a well-founded, interdisciplinary theory of human aggressive and predatory behavior is needed. In this paper, we present a taxonomy of harassment categories to characterize different types of hateful and abusive rhetoric that is common in online social media platforms.  This taxonomy will facilitate the development of harassment filters, the focus of our current research presented here.

\begin{figure}[t]
\centering
\includegraphics[width=0.9\linewidth]{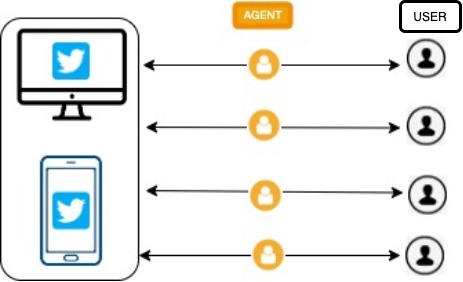}
\caption{Individualized Agent between Social media platform and user helping to customize the filtering of communication to remove or flag online communication.}
\label{fig:Indi}
\end{figure}

The goal of this paper is to develop agent-based user-adapted harassment filters (as shown in Figure~\ref{fig:Indi}) that continually adapts to individual users' threat perception, tolerance, and sensitivity.
As users may have varying sensibility towards different types of harassment, our first goal is to understand the need for 
individualized filtering of harassing tweets. 
To gather ground truth data necessary to identify varying harassment perceptions and to demonstrate the feasibility of training user-adapted filters, we needed user-labeled datasets. To obtain this data, we first collected a set of 
tweets 
using the Twitter API and matching keywords associated with the categories commonly associated with various harassment categories in the taxonomy.  Thereafter, we used a crowdsourcing service, {\tt Amazon Mechanical Turk}, where {\tt MTurk} workers were tasked to label presented tweets based on their perceived harassment intensities and if they would want to filter particular tweets from their Twitter feed 
(see Section~\ref{surveydata} for further details).
The goal of the survey was to understand the variation of sensibility among the users, the variation in the tolerance or acceptance of harassment categories in our taxonomy, and to build agent-based user-adapted filters.
%
%
The collected data is used to answer the following research questions:\\
(1) How does user sensibility vary from category to category?\\
(2) How do the perceptions and acceptances of different harassment types vary in a population?\\
(3) How can the same tweet have different impacts on users, and how do their reactions vary?\\
We performed an in-depth analysis of the collected Survey Data to identify the variation of harassment sensitivity and tolerance levels among the users over different harassment categories in our taxonomy. Our analysis shows how users' perception of harassment intensity and filtering preference varies depending on harassment categories. Also, different users' perceived intensity and acceptance can vary significantly for the same tweet. We evaluated filtering mechanisms for each user and found that agent-based, user-adapted filters are more accurate in predicting user preferences when compared with a general filter trained on the entire dataset. The observations support the need for and benefit of user-adapted filtering of harassing communication.

This paper makes the following contributions:
\begin{enumerate}
  
  \item A taxonomy of harassment categories grounded in psychological literature.
  \item 	A large-scale crowdsourced study (26,000+ responses) analyzing variability in harassment perception.
  \item Statistical validation (ANOVA, Tukey HSD, Wilcoxon) of perception variability.
  \item An agent-based user-adaptive filtering architecture.
  \item Empirical comparison of personalized, general, and majority baseline filters.
    \end{enumerate}

\section{Categories of Cyber-harassment}
\label{sec:Categories}
Most cyber-harassment research 
using binary classification of harassment: a communication either contains harassing content or does not~\cite{chen2012detecting,warner2012detecting,djuric2015hate}. In reality, however, harassment is multifaceted and has many dimensions.
%

Though there is no commonly accepted standard definition of online harassment, most definitions include the following key components: (a) unwanted behavior that occurs through electronically mediated communication, and (b) behavior that violates the dignity of a person by creating a hostile, degrading, or offensive environment~\cite{bossler2012predicting}. Such behaviors can include offensive name-calling, attempts to embarrass, physical threats, stalking~\cite{duggan2014online}, gender harassment, unwanted sexual attention, sexual coercion,
denigration (sending harmful or cruel statement about a person to other people online), impersonation (pretending to be another person in order to make that person look bad), flaming (sending angry, vulgar or rude messages about an individual through an online, public forum), and exclusion (the exclusion of an individual from an online group)~\cite{staude2012stressful}. Definitions vary in terms of whether the behavior must occur multiple times~\cite{lindsay2016personality}, intentionally cause harm~\cite{gidro2016aspects},
and/or involve a perpetrator known to the victim~\cite{duggan2014online}. 

Researchers~\cite{Newman2016online} have emphasized the need for clear characterization and organization of these categories for the field to progress. Much of the current literature examining offline harassment requires knowledge of a perpetrator's intentions, which, while difficult to discern in offline environments, is almost impossible to confirm in online environments such as social media platforms. For this reason, online harassment, as currently understood by researchers, is defined in terms of a victim or third party's understanding rather than a perpetrator's motives. 
Researchers believe there are three main reasons perpetrators may purposefully engage in harassing behavior. 
Purposefully harassing behavior includes (1) rude comments used as a form of self-expression~\cite{coe2014online}, (2) intimidation strategically designed to (a) interrupt communication on a topic or (b) retaliate for past reports or comments~\cite{bernstein2014abuse}; and (3) acts with no strategic aims other than causing psychological or physical harm~\cite{buckels2014trolls}. 
Further complicating an understanding of harassment is the variability in how harassment is perceived. Many definitions require the victim to view the behavior as offensive or threatening~\cite{gidro2016aspects}. Understanding a victim's reaction can be as difficult as discerning a perpetrator's motive when researchers are unable to directly communicate with the victim. 

To understand different types of online harassment, a taxonomy containing different categories of cyber-harassment was prepared and was paired with an associated vocabulary. 
The initial category list is generated from existing literature~\cite{staude2012stressful,page2015only,duggan2014online}: (i) Insults and name calling, (ii) Sending harmful or cruel statement about one user to others, (iii) Religious/racial/ethnic epitaphs, (iv) Sexual orientation, (v) Sex/ gender, (vi) Threat, (vii) Multiple type (message contain more than one harassing type),
(viii) Revealing personal information online about the target, (ix) Tapping, computer viruses, and other digital security threats, (x) General threat of physical harm to self or others, (xi) Specific threats of physical harm to the individual, and (xii) Online impersonation. 
After analyzing a stream of tweets over a period of time, we observed that some of the categories that we initially listed did not appear on Twitter streams. 
Accordingly, we pared the initial list down to the following that was used in our subsequent data gathering and experimentation: 
\begin{description}
\item[General harassment:] Communication is harassing, but does not naturally fit into any other identified category. 
\item[Cruel statement:] Communication contains information that is negative and personal. This information is directed at a specific target and does not specifically address a person's religion, race, ethnicity, sexual orientation, or gender/sex.
\item[Religious/racial/ethnic slurs:] Communication designed to disparage or attack a person's race, religion, or ethnicity.
\item[Harassment based on sexual orientation:] Communication designed to disparage or attack a person's sexual orientation.
\item[Sexual harassment(Gender based harassment):] Communication contains a short, negative label designed to disparage or attack a person's sex or gender. 
\item[Threats of physical harm/violence:] Communication contains a direct threat, metaphorical or actual, against a person's property, family, self, or online presence. 
\item[Multiple types:] Communication contains more than one category of identified harassment. 
\item[Non-harassment:] Communication does not fit into any of the identified harassment categories.
\end{description}
We also identified keywords for each of the identified harassment categories based on frequent and representative words used in corresponding tweets. 

\section{Harassment Data set}
\label{sec:DataSet}
We chose the Twitter platform for collecting harassing communication. Using Twitter's streaming API, we obtained a set of 8000 tweets matching the keywords associated with the initially identified categories and stored them in a MySQL database. As is the case with many real-world datasets, the collected data had several issues that required cleaning and pre-processing to facilitate further research and analysis. After pre-processing, 5231 out of 8000 collected tweets were found to be usable.  
For each tweet in this set, we had three individuals (undergraduate and graduate students in our lab) label it according to one of the category types in our taxonomy. The majority label was then associated with the tweet and used for our experiments.
The result of the labeling is presented in Table~\ref{tab:1}, which contains the number of tweets assigned to each of the aforementioned categories. We refer to this dataset as the {\tt Twitter Categorical dataset}.
 Almost 45\% of this usable data is categorized as non-harassing or does not fit into any identified categories. 
We observed that several data points lacked unanimous labels. This observation shows that there is a pronounced variation in harassment perception among the human labelers. 

\begin{table}
\centering
\caption{Frequency of Tweet Categories}
\label{tab:1}
\begin{small}
\begin{tabular}{||c|c|c||}\hline\hline
\# &Category & \# of tweets  \\\hline \hline 
0 & General harassment& 79\\\hline 
1 & Cruel statement & 1054 \\\hline 
2 &  Religious/racial/ethnic &89 \\\hline 
3 & Sexual orientation &11 \\\hline 
4 &  Sex/ gender &656 \\\hline 
5 &  Threat & 236\\\hline 
6&  Multiple types & 106 \\\hline 
7 &  Non-harassment & 2382\\\hline
N/A & Non-codable & 618 \\\hline \hline
 & Total data & 5231\\\hline\hline
\end{tabular}
\end{small}
\end{table}
\section{Survey Data: User-Sensitivity Data}
\label{surveydata}
We wanted to understand how different users have distinct sensibilities towards different harassment categories.
We used Amazon Mechanical Turk~\cite{turk2012amazon} (MTurk) workers to rate harassment intensities and the need for filtering of individual tweets.
The MTurk survey participants were provided a collection of unlabeled tweets, drawn from 
multiple harassment categories from the Twitter Harassment dataset. 
Participants were asked to respond to the following two questions about each tweet:\\
(1) What intensity of harassment is present in the following tweet?\\Options: {\em (a) None, (b) Minimal, (c) Moderate, (d) High, (e) Extreme.}\\
(2) Do you want to filter this tweet?\\Options: {\em (a) Yes, (b) No.}

The survey is prepared using Django~\cite{holovaty2009definitive}, a web application based on Django, hosted on the Heroku website~\cite{middleton2013heroku}. 
For each MTurk participant, the survey contained 75 tweets selected from all categories of the Twitter Categorical data. Each tweet was presented to 5 MTurk workers. Around 360 MTurkers successfully completed this study 
over a two week period
which allowed us to collect around 26,500 responses. We used this dataset to understand how perception and acceptance of the different type of harassment vary in a population.  As described below, the analysis of the dataset also provided the rationale for user-adapted, agent-based harassment filters learnt from user filtering choices.

\section{Need for Individualized Learning Agents for Filtering Tweets}
We wanted to understand if users need personalized agents trained for customized filtering of harassing Tweets based on individual preference, or if a general filter, trained on the entire data set serve the purpose. To determine whether the individualized agent-based harassment filters are warranted, we tried to understand if different users have distinct harassment perception and filtering sensibility/preference.
To assess user sensibilities towards different categories and intensities of harassing communication, we presented subsets of tweets selected from the Twitter harassment dataset to MTurk workers, where workers were asked to rate, for each tweet presented, their perceived harassment intensity and if they wanted that tweet to be filtered from their input stream.

\begin{figure}[t]
\centering
\includegraphics[width=0.9\linewidth]{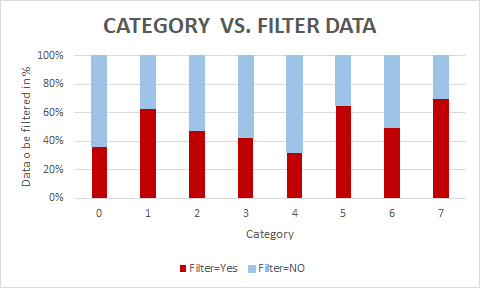}
\caption{Percentage of data selected by users to be filtered for the different harassment categories.}
\label{fig:filter}
\end{figure}

The percentage of data in each of the harassment categories that the entire group of MTurk users participating in the study wanted filtered is shown in Figure~\ref{fig:filter}. 
The 'red' colored part of the bar represents the percentage of tweets to be filtered for that category. We observe that there are significant variations in the percentage of data to be filtered for different categories.
\begin{observation}
User filtering preferences or acceptance of cyber-harassment vary significantly by harassment categories.
\end{observation} 

\begin{figure}[t]
\centering
\includegraphics[width=0.9\linewidth]{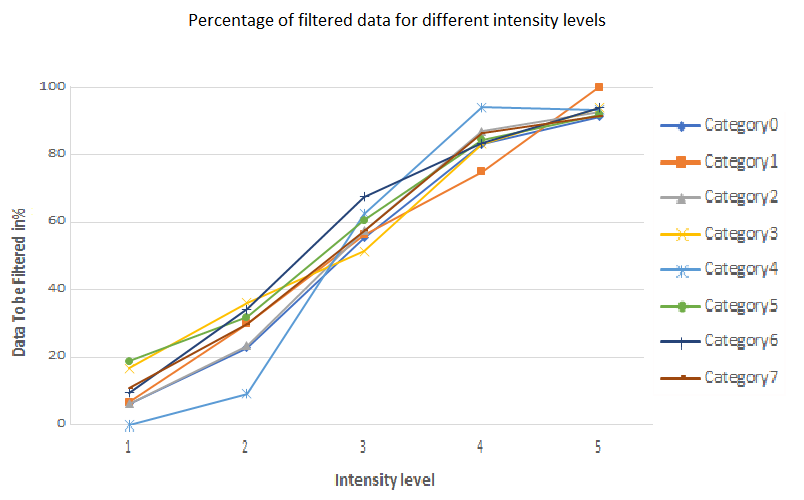}
\caption{Percentage of tweets selected to be filtered for each category and for 
perceived harassment intensity levels. 
}
\label{fig:filIntenisty}
\end{figure}

Figure~\ref{fig:filIntenisty} depicts the percentage of tweets to be filtered for each category given the perceived harassment intensity. In the figure, the X-axis represents perceived intensity levels. The Y-axis represents the percentage of tweets selected to be filtered. 
\begin{observation}
The percentage of tweets selected for filtering increases with the increase in perceived harassment intensity. 
\end{observation}

This pattern holds for each of the harassment categories. This observation supports the hypothesis that users have lower acceptance of higher intensity of cyber-harassment.  The figure also shows that the tolerance level varies between categories for the same harassment intensity.  That leads to the following critical observation:
\begin{observation}
Harassment tolerance depends both on the category of harassment and the perceived intensity of harassment.
\end{observation}

\subsection{Intensity level influence on filtering choice}
In this section, we present results from statistical tests to determine the significance of the observed trends in the survey.  
\subsubsection{ANOVA test:} We used the one-way ANOVA test~\cite{eriksson2008cv} 
to determine if the choice to filter tweets varied significantly by the perceived harassment intensity level. ANOVA checks the impact of one or more factors by comparing the means of different samples. Using this test, we can conclude if there is a statistically significant difference between the percentage of tweets selected for filtering between different perceived harassment intensity levels. 
The following two hypothesis are considered in ANOVA testing:\\
{\bf Null hypothesis (H0):} All intensity levels have a similar effect on users.\\
{\bf Alternate hypothesis:} Different intensity levels have significantly different effect on users.\\
The ANOVA test for the collected data for different intensity levels yielded an F-statistic value that measures the differences in the means of different intensity levels and suggests whether the levels are significantly different or not. $\alpha$ is considered the significance level and the probability of rejecting the null hypothesis when it is true. If the F-statistic is more than the F-critical value for the chosen $\alpha$ value, then the null hypothesis (H0) can be rejected, and we can say that different intensity levels have significant effects on users. 
For the data presented above, the ANOVA test returned an F-value of 304.7577, which is greater than the F-critical value, 2.64146, for the selected alpha level of 0.05. 
The ANOVA result is $ [F(4,35)=304.758, p=1.139e^{-26} <0.05]$.
Hence the null hypothesis (H0) can be rejected, which implies that different harassment intensity levels have a significantly different effect on user filtering decisions.

In addition to checking the statistical significance, it is useful to report the effect size measure for an ANOVA test, which reflects the size of the performance difference of the alternatives being considered. 
A high effect size value signifies that not only are there more than two groups that significantly differ from each other, but also suggests that the difference is significantly high. For the ANOVA test we ran, the effect size value, $\eta^2=\frac{SS_{intensities}}{SS_{Total}}$, is 0.972. This suggests that some of the intensity levels are different from others by a high margin.

The limitations of the one-way ANOVA testing is that while it can suggest that at least two of the tested data groups were different from each other, it cannot identify which groups were different. 
To identify those groups, we needed to further test what are the intensity levels that result in significant differences. 
For this purpose we use the Tukey test, which is a statistical significance test that is often used to identify the effect size with ANOVA analysis.
\subsubsection{Tukey Honest Significant Difference (HSD):} Tukey's Honest Significant Difference (HSD) test 
The Tukey test is a post-hoc test based on the studentized range distribution. 
Tukey's HSD is considered to be the strongest test to find out which groups are significantly different and is widely used. This test calculates all possible pairs of means and then calculate the significant difference. 
Based on the computed q-statistics by the Tukey test on our dataset, we observe that each of the intensity levels 1, 2, and 3 are significantly different from all the other intensity levels with a 99\% confidence. Intensity 4 is different from intensity 5 with 95\% confidence value. 

\subsection{Harassment category influence on filtering choice}
We have so far presented statistical analysis to understand whether users' filtering choices are affected by the perceived intensity level. In this section, we discuss the statistical significance of the type of harassment contained in the received communication on the user's filtering choice.
In particular, for a given perceived harassment intensity level, does the percentage of tweets marked for filtering vary for different harassment categories?
As we have only one value, namely the percentage of tweets marked for filtering for each harassment category, for each intensity level, we needed a different significance testing mechanism than ANOVA.  In the following, we present the results from using the Confidence Interval Coefficient measure used for this analysis. 

{\em Confidence Interval Coefficient}~\cite{agresti2008simultaneous} compares proportions of samples through constructing simultaneous confidence intervals and P-difference for the confidence intervals. We constructed confidence intervals to make pairwise comparisons of the percentage of tweets to be filtered for different categories for a given intensity. The significance is determined by comparing the {\em p-difference} with the $\alpha$ value (we chose $\alpha=0.05$). If the p-difference calculated for confidence intervals is greater than the $\alpha$ level selected, it suggests that for the same intensity level, there is a significant difference in the chosen categories. 
Results of Confidence Interval Coefficient for intensity level '2', corresponding to the data presented in Figure~\ref{fig:filIntenisty}, is shown in Table~\ref{result:confidence2}. 
The rows that are highlighted in green indicate the categories which are significantly different from each other for the given intensity level. Results in Table~\ref{result:confidence2} show that for intensity level '2', users choose to filter a statistically significantly different percentage of tweets for category 0 compared to category 3-7.
Tables for other perceived harassment intensities cannot be accommodated due to space constraints.  We have observed that user acceptance of different harassment categories vary significantly at lower intensity levels. As the perceived intensity increases, however, the differences in the filtering choices of harassment of different categories were eliminated. This is primarily due to the fact that more and more tweets, irrespective of categories, are selected to be filtered by users at high perceived harassment intensities.
\begin{table}
\caption{Confidence Interval Coefficient for \% tweets filtered for Intensity Level 2 (columns highlighted denote statistically significant difference; 19 of 28 pairs are different).}
\label{result:confidence2}
\begin{center}
\begin{small}

	 \begin{tabular}{||c|c|c|c|c||}
      \hline\hline
	      Category & Category& P-Diff &  Lower & Upper\\
          \hline\hline
          \rowcolor{Green}
          	 0 & 1&  0.0704 &-0.2139 & 0.0759\\
          \hline
      		 0 & 2& 0.0055& -0.1448 &0.1339\\
          \hline
          \rowcolor{Green}
     		 0 & 3&  0.133 &-0.2791 &0.0182\\
          \hline
          \rowcolor{Green}
 			 0&  4&  0.137& 0.01433 &0.2546\\
          \hline
          \rowcolor{Green}
 			 0& 5&0.0898& -0.2342& 0.0582\\
          \hline
          \rowcolor{Green}
 			 0& 6&0.1142& -0.2597& 0.0357\\
          \hline
          \rowcolor{Green}
 			 0& 7&0.0698& -0.2133& 0.0764\\
          \hline
          \rowcolor{Green}
			 1 & 2&  0.0649& -0.0818& 0.2091\\
          \hline
          \rowcolor{Green}
			 1 &3&0.0626 &-0.2157& 0.0929\\
          \hline
          \rowcolor{Green}
			 1 &4&0.2075 &0.0764& 0.3307\\
          \hline
			 1 &5& 0.0194&-0.1709& 0.133\\
          \hline
			 1 &6& 0.0438& -0.1963& 0.1105\\
          \hline
			 1 &7&0.0006 &-0.15007 &0.1513\\
          \hline
          \rowcolor{Green}
 			 2&3& 0.1275& -0.2742& 0.02415\\
          \hline
          \rowcolor{Green}
 			 2&4& 0.1427 &0.01909&0.2607\\
          \hline
          \rowcolor{Green}
 			 2&5&0.0843& -0.2294& 0.0641\\
          \hline
          \rowcolor{Green}
			 2&6& 0.1087 &-0.2548& 0.0416\\
          \hline
          \rowcolor{Green}
			 2&7& 0.0632& -0.0923& 0.2163\\
          \hline
          \rowcolor{Green}
			  3&4&   0.2702& 0.1335 &0.3963\\
          \hline
			3&5& 0.0432 &-0.1132 &0.1979\\
          \hline
			 3&6&0.0188& -0.1385& 0.1754\\
          \hline
          \rowcolor{Green}
			3&7& 0.0632& -0.0923& 0.2163\\
          \hline
          \rowcolor{Green}
			 4& 5& 0.2269 &-0.3511 &-0.0938\\
          \hline
          \rowcolor{Green}
			 4&6& 0.2514& -0.3767& -0.1162\\
          \hline
          \rowcolor{Green}
			 4&7&0.2069& -0.3299 &-0.0758\\
          \hline
			 5&6&0.0244& -0.1786& 0.13073\\
          \hline
			 5&7 &0.01999& -0.1324& 0.17157\\
          \hline
			 6&7&0.0444& -0.1098& 0.1969\\
          \hline\hline
    \end{tabular}
    \end{small}
\end{center}
\end{table}

To summarize, statistical tests on the user responses confirms that user's choice of tweets to filter is dependent on both the category and the intensity level of a particular tweet.
Results from ANOVA testing and Tukey test shows that user filtering choices have significant differences with varying intensity levels. The corresponding effect sizes were also found to be high.  Results of the Confidence Interval Coefficient show that user reaction varies for many categories of harassment for the same intensity levels.  These observations highlight the need for user-adapted harassment filters. 

\section{Individualized Agent-Based Harassment Filters} 
We now focus on the development of agent-based user-adapted filtering of harassment content for the users. 
As observed above, user tolerance for harassing communication varies based on intensity level and harassment category.  In addition, different users have different filtering preferences for different intensity levels and categories. So a general filter, trained on labeled data from all users, will be unable to meet the filtering needs of individual users. 
\begin{figure}[t]
\centering
\includegraphics[width=0.9\linewidth]{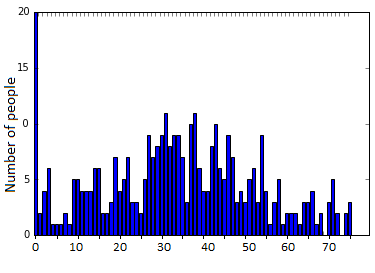}
\caption{Histogram of \# users wanting to filter a certain number of tweets. X axis represents \# of tweets (0-75), Y axis represents \# of users wanting to filter that many tweets.
}
\label{fig:hist}
\end{figure}

To develop an adaptive personal agent for filtering harassing communication,
We performed further analysis of the Survey Data. 
In Figure~\ref{fig:hist}, we present a histogram of the counts for users with the number of tweets they wanted to be filtered.  Each bar in the histogram shows how many people wanted to filter that number of tweets out of the 75 tweets they labeled. 
The histogram demonstrates that there is considerable variation in the population based on individual users' acceptance of harassment and presents a compelling case for the need for user-adapted filtering of harassing communication. The results also clearly demonstrate a wide variety in user tolerance levels.  On one extreme, participants were extremely tolerant about harassing content and chose not to filter any of the tweets. On the other extreme,  participants were very sensitive to harassment and chose to filter all the tweets they received.

\begin{figure}[t]
\centering
\includegraphics[width=0.85\linewidth]{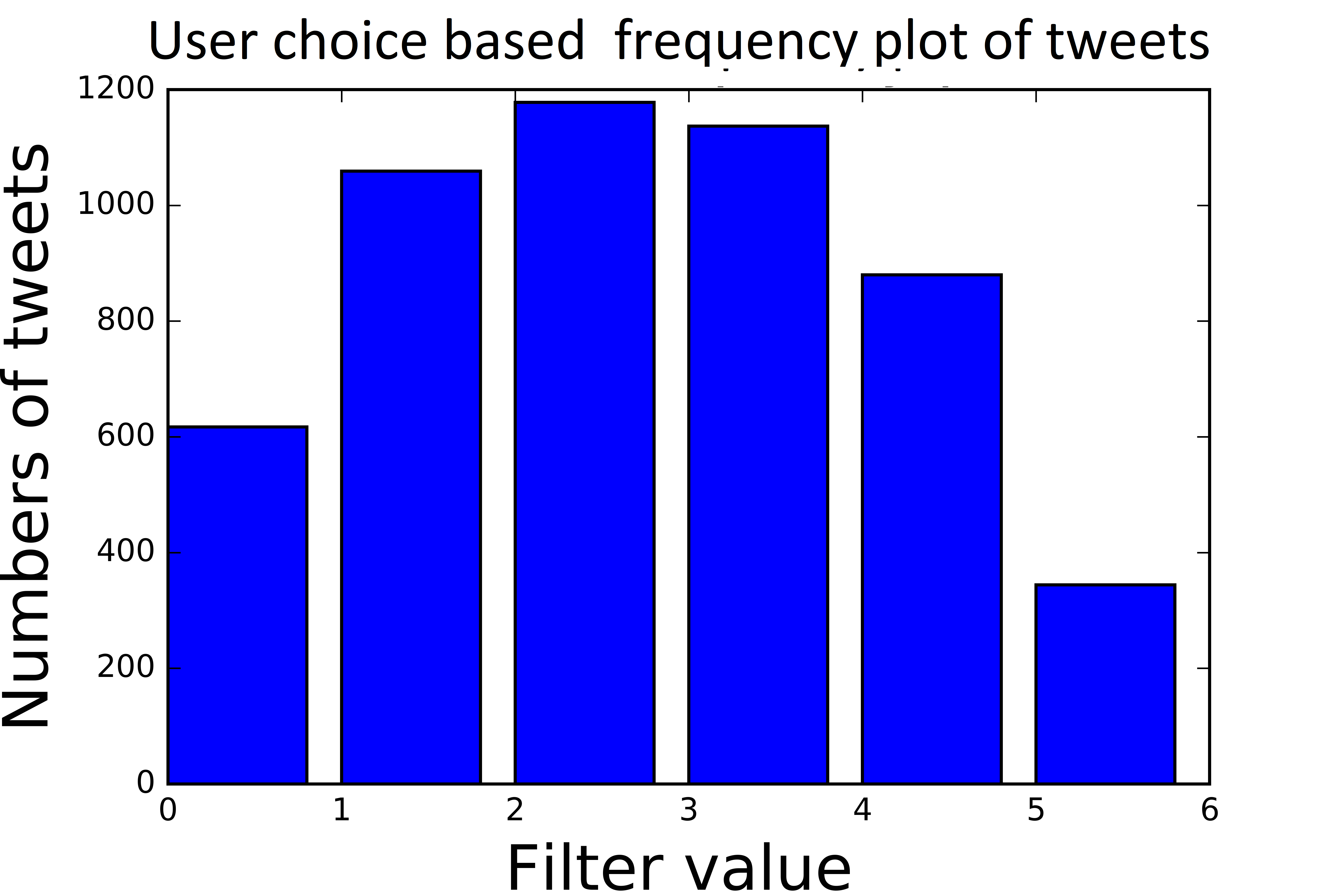}\\ (a)\\
\includegraphics[width=0.85\linewidth]{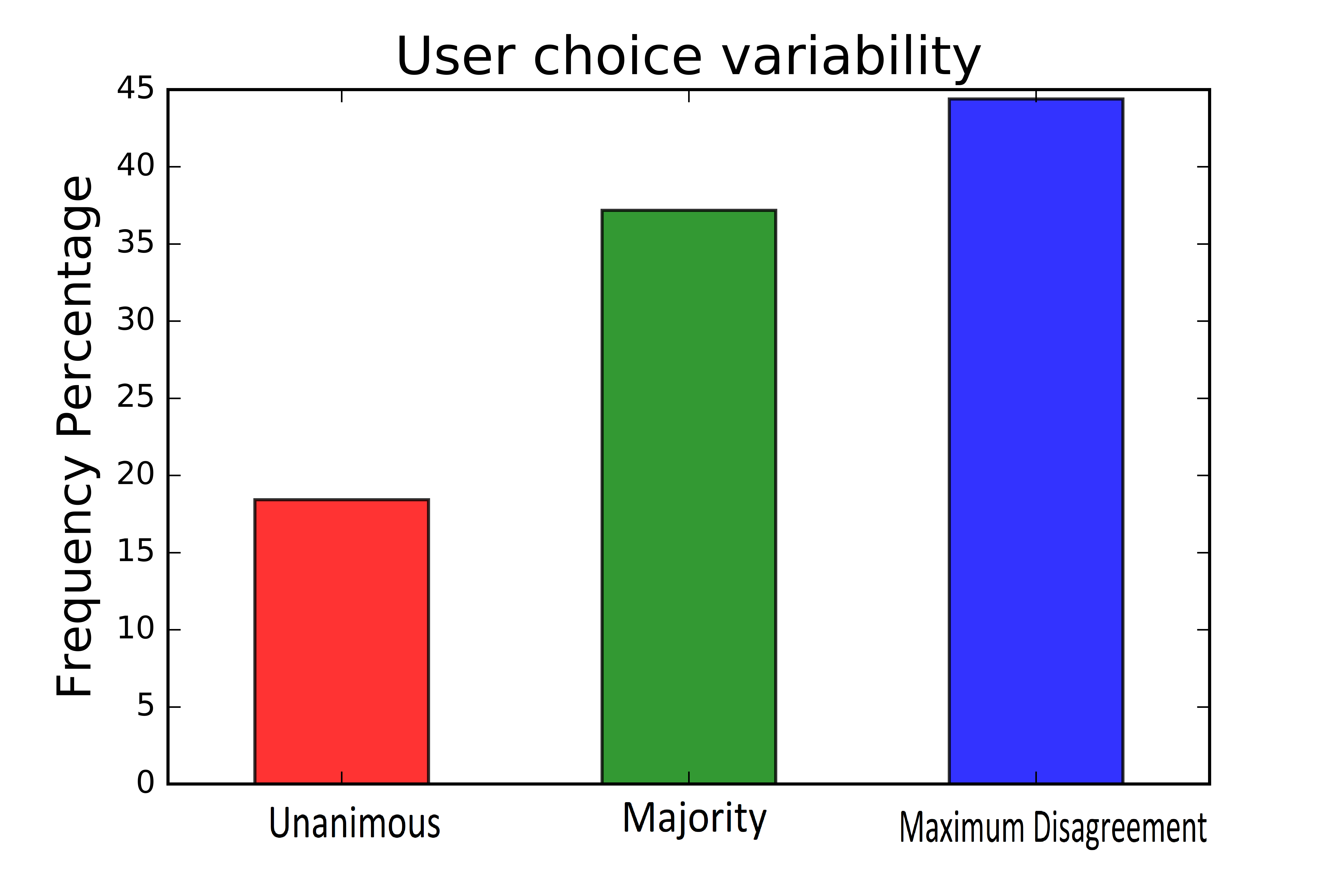} \\ (b)
\caption{(a) Frequency plot for filtering choices
(X-axis: \# users, out of 5, who chose to filter a tweet, Y-axis: \# of tweets), (b) 
User Choice variability plot.}

\label{fig:dif_intensity}
\end{figure}


We recall that each tweet from the Twitter Categorical data was labeled by 5 MTurk users for harassment intensity and filtering preference.
In Figure~\ref{fig:dif_intensity}(a), we present a count of tweets that received a certain number of votes, from 0 to 5, for filtering.
In Figure~\ref{fig:dif_intensity}(b), we show how many tweets were unanimously voted by all the five users to filter/not-filter, 
How many got significant \emph{majority} (4/1 or 1/4 counts) of filter/not-filter selection, and how many tweets had \emph{maximal disagreement} (3/2 or 2/3 counts) for filter/no-filter choices. Results show that only 18\% of the tweets were unanimously voted on for filter/no-filter. Approximately 38\% and 45\% of tweets were marked for majority and maximal disagreement groups, respectively. These observations clearly demonstrate that users vary considerably in their harassment sensibility and acceptance.

\subsection{A general harassment filter}
We first created a general filter that learns to filter 
based on the majority of the filter/no-filter value for each tweet. The filter-label of each tweet is decided based on the majority of the 5 votes from users who labeled that tweet. Developing a single filter for all users, rather than user-adapted, agent-based filters learnt from interactions with each user, is computationally cheap. We believe, however, that a single general filter will not be effective for all users because of significant false positives and false negatives, and hence precision and recall errors.  In particular, such a general filter is unlikely to be able to protect a sensitive user from exposure to 
unacceptable harassing tweets. 
\begin{figure}
\centering
\includegraphics[width=0.3\textwidth]{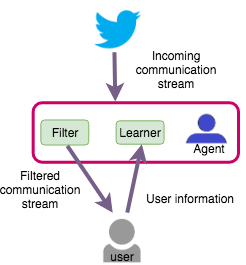}
\caption{An agent-based user-adapted harassment filter.}
\label{fig:agent}
\end{figure}
 
\begin{figure}
\centering
\includegraphics[width=0.49\textwidth]{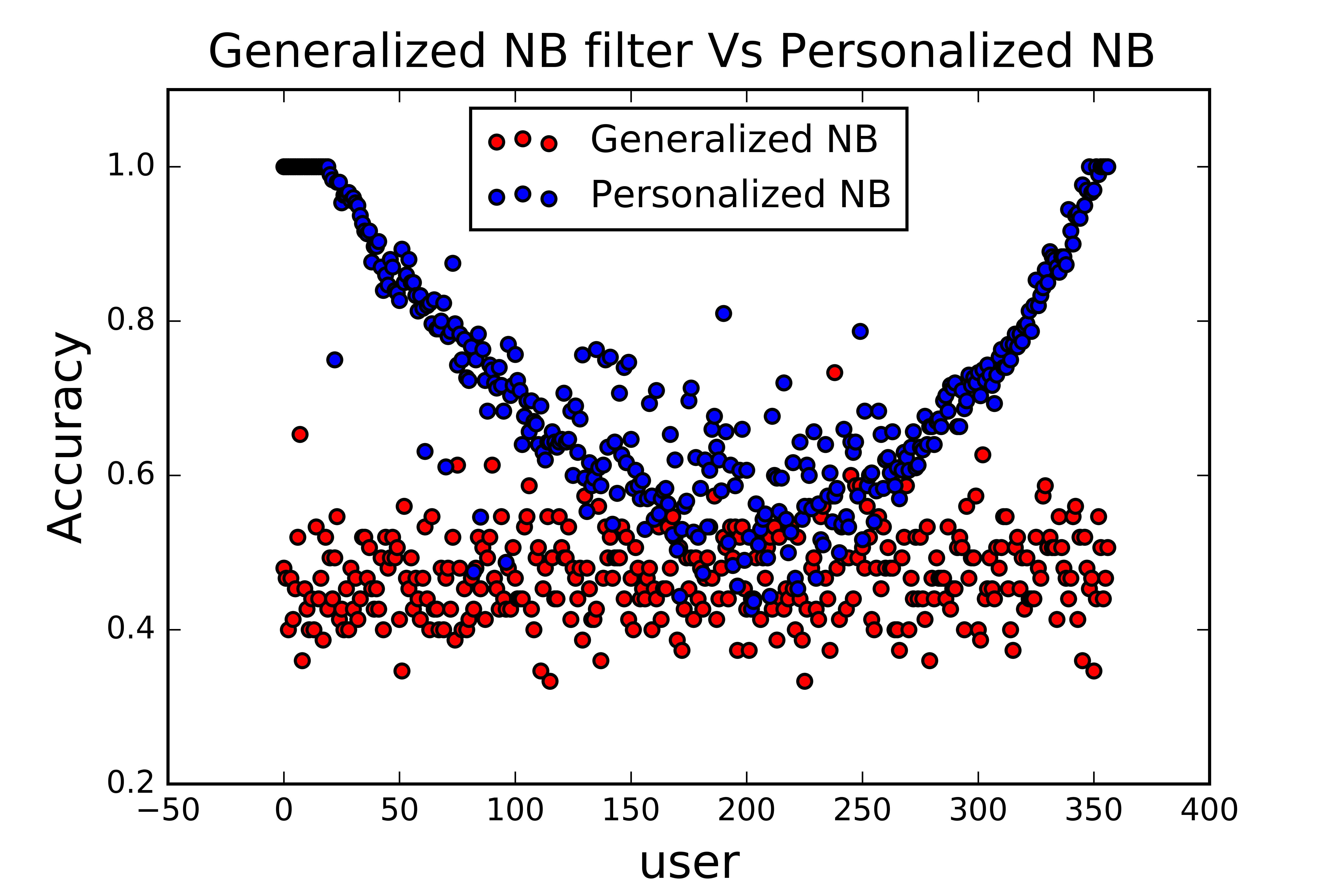}
\caption{Accuracy of General Vs user-adapted  Filters.} 
\label{fig:NB}
\end{figure}

\subsection{Agent-based Individualized filters}
In the agent-based filtering approach we propose, agents are not only initially trained by learning from available user filtering choices, but can also continuously monitor and adapt based on the users' online behavior. 
Agents can therefore adapt their filtering behavior to accurately reflect changing user sensibilities and protect them from harassing encounters on their social media accounts.
In our experiments, we allocated one agent for each user in the system (Figure~\ref{fig:agent} presents the interaction between the agent and its associated user). We provide training data from the filtering choices on tweets to an agent from its associated user. That agent then used effective supervised classification algorithms to learn user-adapted filters for its user. 
While this modality of agent-based filter development requires much more effort, as one agent has to be trained for each user, we believe that performance improvements in intervention by shielding users from unwanted harassing communication justify this additional computational cost.  

We used Naive Bayes (NB)~\cite{mccallum1998comparison}, Support Vector Machine (SVM)~\cite{hearst1998support}, and Random Forest (RF)~\cite{liaw2002classification} supervised learners to train user-adapted filters. We also used the same classification mechanisms to build the general filters. In Figure~\ref{fig:NB}, we present a comparison of the 10-fold cross-validation accuracy of a Naive Bayes based user-adapted filter for each agent to that of the corresponding general filter when used for each user.  
In Figure~\ref{fig:NB}, the X-axis represent each user and the Y-axis represents the prediction accuracy of each user's filtering choice using the general and agent-based user-adapted filtering mechanism.  
The users are sorted in non-decreasing order of the number of tweets they chose to filter out of the 75 presented to them, i.e., higher numbered users chose to filter a higher percentage of the tweets they rated.

We observe that for almost all the cases, user-adapted filter significantly outperform the general filter.  The V-shape of the accuracy plot for the user-adapted filter reflects the fact that it is easier to predict the choice of users who chose to filter almost all or almost none of the tweets, compared to more discerning users, e.g., chose to filter half of the tweets rated.  We also find the accuracy rate of predicting the choice of the more discerning users to be more dispersed: choices of some of these users are significantly easier to predict than others.  While the accuracy rates of filtering choices for the users in the middle range are not stellar, an online agent may be able improve its prediction with more interactions with the associated user.  There are also a few odd instances where the general classifier do well; we believe those users may have filtering preferences more aligned with the "population average". 
Trends from the user-adapted and general filter trained using SVMs and Random Forest classifiers are similar to the NB classifiers.

\subsection{Comparison of agent-based  classifier with majority filtering decision}
\begin{figure}[t]
\centering
\includegraphics[width=0.49\textwidth]{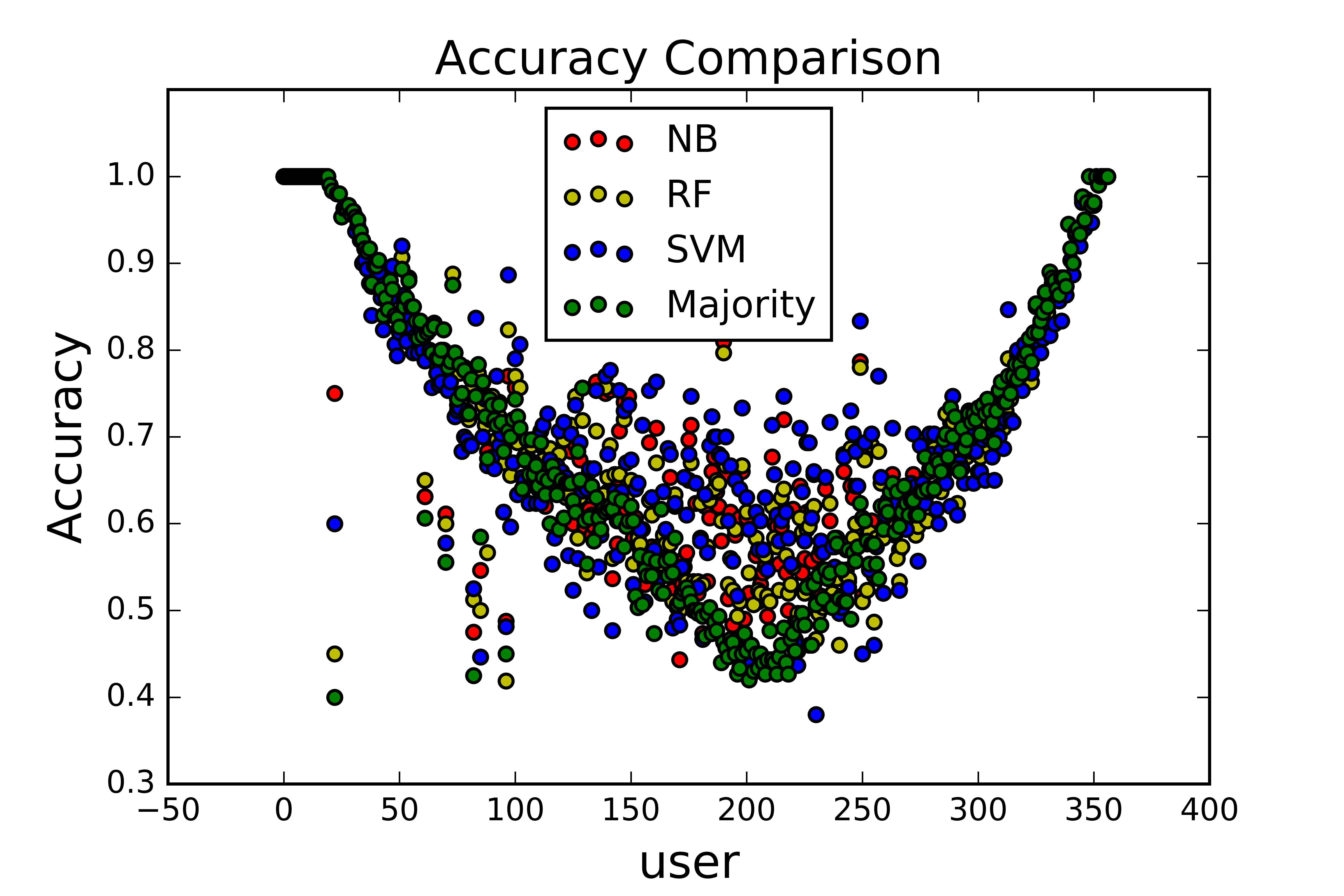}
\caption{Majority class Vs 
NB, SVM, RF classifiers.} 
\label{fig:comparison}
\end{figure}

We did further analysis with a Majority Class filter, which uses the majority filtering decision of a user, i.e., if a user chooses to filter a majority of the tweets presented then all tweets are filtered and vice versa. The purpose of using the Majority filter is to check whether an agent is able to leverage user-specific training data to build a user-adapted filtering mechanism and improve on a baseline filtering scheme, which is also based only on that user's choices (in contrast to the general filter in the previous section that is trained on data from all users).
In Figure~\ref{fig:comparison}, the X-axis represents each user, ordered in increasing number of the tweets they chose to filter, and the Y-axis represents the accuracy of each user for a user-adapted  Majority Class filter as well as all three (NB, SVM, and RF) agent-based filters.  
Table~\ref{table:algocomp} shows the number cases the majority classifier did better (win), worse, or tied with each of the learning filters.
Results show that the user-adapted  classifiers using Naive Bayes, 
Random Forest and SVM classifiers perform slightly better than the user-adapted majority filter, whereas the Naive Bayes filter significantly outperforms it. 

To further verify if the performance differences of the user-adapted learning and the majority filters are statistically significant, we use the Wilcoxon Signed Rank Test~\cite{woolson2007wilcoxon}, which gives the
following results: (a) the user-adapted filtering mechanisms based on Naive Bayes, Support Vector Machine, and Random Forest performed significantly differently from the user-adapted  Majority class classifier, (b) the Naive Bayes filter performed significantly different than Random Forest and Support Vector Machine, (c) the performance of Random Forest and Support Vector Machines is statistically indistinguishable.
These results show that the agent-based filtering mechanisms are able to leverage user-specific training data to improve on a user-adapted baseline classifier.

\begin{table}
\caption{Comparison of performance of the learning algorithm with the majority Algorithm.}
\label{table:algocomp}
\begin{center}
\begin{small}
	 \begin{tabular}{||c|c|c|c||}
      \hline
	    Majority Vs & SVM & RF & NB \\ 
	     \hline\hline
	    Win & 149 & 132 &52 \\
	     \hline
	   Lose  & 166 & 143 & 136   \\
	     \hline
	     tie  & 42 & 83 &  169 \\
	     \hline\hline
    \end{tabular}
    \end{small}
\end{center}
\end{table}

\section{Related Work}
\label{sec:background}
Prior work on harassment detection spans 
several web platforms including Twitter~\cite{burnap2014hate}, Facebook~\cite{rybnicek2013facebook}, Instagram~\cite{hosseinmardi2015analyzing}, Yahoo!~\cite{nobata2016abusive}, YouTube~\cite{dinakar2012common}, 
, etc. 
Different platforms have unique communication modalities, user demographics, and content and may, therefore, display different subtypes of hateful communication. For instance, one should expect quite different types of hate content on a platform catering to adolescents than on a web platform used by a wider cross-section of the general public.
Manual analysis of data and the establishment of relationships between multiple features are often error-prone.
Machine learning has been used to address this issue.

In ~\cite{yin2009detection}, a supervised classification technique is used along with local, sentimental, and contextual features extracted from a post using Term Frequency-Inverse Document Frequency (TF-IDF). The classification technique is conjugated with n-grams and other features, such as incorporating abusiveness, to train a model for detecting harassment. A significant improvement over the general TF-IDF scheme is observed 
while adding the sentimental and contextual features.
In~\cite{sood2012using}, Support vector machines (SVMs) were used to learn a model of profanity using the bag of words (BOW) approach to find the optimal features.
This approach surpasses the performance of all previous list-based profanity detection techniques.
%
%
A linguistic and behavioral pattern-based model~\cite{mosquera2014detecting} was proposed to filter short texts, detect spam and abusive users in the network. It used a real-world SMS data set from a large telecommunications operator from the US and a social media corpus. It also addressed different ways to deal with short text message challenges, such as tokenization and entity detection by using text normalization and substring clustering techniques.
%
%
A comprehensive approach to detecting hate speech was proposed in~\cite{warner2012detecting}, which presents a plan that targets specific group characteristics, including ethnic origin, religion, gender, and sexual orientation.
The paragraph2vec 
approach~\cite{djuric2015hate} is used to classify anti-Semitic speech on data collected over a 6-month period from Yahoo Finance website.
In~\cite{nobata2016abusive}, a comprehensive lists of slurs, obtained from hate speech and an array of features for abusive language detection. such as POS tags, the presence of blacklisted words, n-gram features including token and character n-grams and length features, are used. Their scheme outperformed a deep learning approach by focusing on good annotation guidelines that help detect specific abusive language.
In~\cite{burnap2016us}, an exploratory single blended model of cyber-hate that incorporates knowledge of features across multiple types was used. The proposed method improved classification for different types of cyber-hate beyond the use of a BOW and known hateful terms.
In~\cite{waseem2016hateful}, the author analyzed the impact of various extra-linguistic features in conjunction with character n-grams for hate speech detection. It was observed that though the gender feature is informative, differences in the geographic and word-length distribution do not improve performance over character-level features. A list of criteria based on critical race theory to identify racist and sexist slurs was presented.

Most of the research studies that we have come across, including the ones summarized above, have focused on binary classification of communication. Cyber-harassment, however, is multi-faceted and may convey diverse content, including hostility, humiliation, insults, threats, unwanted sexual advances, etc., to a target. 

Human moderation efforts for curbing online abuse have proved to be inadequate, and automated agents can be effective tools for mitigating cyber-harassment. Despite the research on automated harassment detection, quality research on automated intervention is sparse. Two recent studies, however, explore automated intervention: (i) ~\cite{geiger2016bot} uses bots to help maintain block-lists, but requires manual user additions, and (ii) ~\cite{Trollacm2016} uses a bot, imitating a person with similar demographic characteristics to the troll, to provide feedback to dissuade hateful comments. These approaches do not consider individual users' harassment perception or sensibility. In~\cite{sen2018agents}, the author discussed some research challenges that can be approached with an intelligent agent-based solution. The author also sheds light on how a user-based agent 
can be useful to learn the user's threat perception and acceptance, and filter different abusive incoming communications directed towards the user.

As there is a gap between psychological theories and computational models and concepts of what constitutes aggression, training computers for comprehensive harassment detection has proved to be challenging. The goal of this research is to remove this disconnect and leverage computational and psychological approaches to identify different aggression categories, crowdsourced feedback, surveys, and statistical techniques. Based on the identified categories, an effective agent-based detection tool has been built to filter communication based on users' sensibilities and preferences.

In recent years, deep neural architectures and transformer-based language models have substantially advanced the state of the art in abusive language and hate speech detection. Large pre-trained models fine-tuned on platform-specific corpora have demonstrated strong performance in binary and multi-class harassment classification tasks. However, the majority of these approaches optimize global predictive accuracy and rely on population-level labeling, often overlooking individual differences in harassment perception and tolerance. In contrast, our work emphasizes personalization-aware moderation through agent-based, user-adaptive filtering mechanisms. By modeling variability in user sensitivity and incorporating individualized learning, this framework complements modern detection systems and highlights the importance of adaptive intervention in online safety infrastructures.

\section{Conclusions and Future Work}
Our goal is to develop a better understanding of cyber harassment types, the varying perception among users to harassing communication, as well as the variation in tolerance of harassment categories, and hence the need for user-adapted filtering of harassing communication. To achieve our goal, we used a taxonomy of harassment categories grounded in the psychology literature, collected a set of tweets that matched identified keywords associated with different harassment categories, developed a crowd-sourced dataset where MTurk workers were asked to rate the harassment intensity of presented tweets, and if they wanted to filter them from their input stream. We performed an in-depth analysis of the labeled data to identify the variation of harassment sensitivity and tolerance level among the users over different harassment categories.
These results show how user's perspective varies depending on harassment categories and the intensity of the tweets. As different users' perceived intensity and acceptance could vary significantly for the same tweet, this highlighted the need for an adaptive learning agent that can use continuous and user-adapted intervention mechanisms to mitigate the effects of cyber-harassment. 

We first trained a general filter from the entire labeled data set and tested it individually on each user's data. The general filter failed to protect sensitive users from exposure to unacceptable tweets, as it only learns the filtering decision for a tweet from the population majority of filter/no-filter labels. Next, we trained separate agent-based filters for each user from that user's filtering choices (from only $\approx$60 labeled tweets per user) and evaluated their ability to filter harassing communication for the corresponding user. 
The user-adapted filters performed better than the general filter for a large majority of the users.  

We also found that the agent-based user-adapted filters outperformed a baseline non-learning filter using the majority filtering choice for corresponding users. These results reaffirmed the pressing need for user-adapted harassment filtering mechanisms rather than relying on a single, generic filter for all. 

 This work can be extended to provide supportive communication to the victims of cyber harassment/aggression. The victims of aggressive communication can be heartened by receiving supportive communication. A user-specific agent can identify personality traits and the aggression tolerance levels of the user and can provide personalized supportive communication that dilute the chain of harassing comments received, divert attention from these negative inputs, and bolster the confidence and self-esteem of the target of aggression. A complementary promising future research avenue for developing effective interventions would be to develop an agent that sends responses to online aggressors with the goal of disarming, discouraging, and eliminating future harassment. Based on the learned user perception and tolerance, an agent can detect aggressive tweets and can identify the source. Based on other attributes like the frequency, intensity level, and the number of users attacked by such sources, an agent can detect potential threat. An agent can then send commensurate responses to the aggressor or harasser to mitigate the situation and disarm the aggressor.
 
 In the future, we can investigate unsupervised grouping of users with similar harassing preference which will allow training of a limited number, one per group, of user-adapted filters. An orthogonal research direction would be for adaptive agents to track changing preferences of the user and accordingly adapt the associated filter.  This is particularly relevant as users change their attitude towards particular categories of harassment based on changing personal life situations and interactions with new acquaintances.


\bibliographystyle{ACM-Reference-Format}  
\bibliography{MyBibliography}
\end{document}